\documentclass[10pt,twocolumn,letterpaper]{article}

\usepackage[pagenumbers]{cvpr}      

\usepackage{booktabs}
\usepackage{makecell}
\usepackage{times}
\usepackage{epsfig}
\usepackage{graphicx}
\usepackage{amsmath}
\usepackage{amssymb}
\usepackage{verbatim}
\usepackage{algorithm}
\usepackage{algpseudocode}
\usepackage{enumitem,kantlipsum}
\usepackage{geometry}
\usepackage{tabularx}
\usepackage{booktabs}
\usepackage{tabularx} 
\usepackage{algpseudocode}
\usepackage{enumitem,kantlipsum}
\usepackage{adjustbox}
\usepackage{tikz}

\usepackage[pagebackref,breaklinks,colorlinks]{hyperref}

\usepackage[capitalize]{cleveref}
\crefname{section}{Sec.}{Secs.}
\Crefname{section}{Section}{Sections}
\Crefname{table}{Table}{Tables}
\crefname{table}{Tab.}{Tabs.}

\def\cvprPaperID{*****} 
\def\confName{CVPR}
\def\confYear{2023}

\begin{document}

\title{Innovative Horizons in Aerial Imagery: LSKNet Meets DiffusionDet for Advanced Object Detection}

\author{Ahmed Sharshar\textsuperscript{*}, Aleksandr Matsun\textsuperscript{*}\\
Mohamed bin Zayed University of Artificial Intelligence (MBZUAI)\\
AbuDhabi, UAE\\
{\tt\small \{ahmed.sharshar, aleksandr.matsun\} @mbzuai.ac.ae}
}
\maketitle

\begin{abstract}
In the realm of aerial image analysis, object detection plays a pivotal role, with significant implications for areas such as remote sensing, urban planning, and disaster management. This study addresses the inherent challenges in this domain, notably the detection of small objects, managing densely packed elements, and accounting for diverse orientations. We present an in-depth evaluation of an object detection model that integrates the Large Selective Kernel Network (LSKNet)~\cite{lsknet} as its backbone with the DiffusionDet~\cite{Diffusiondet}  head, utilizing the iSAID dataset~\cite{isaid} for empirical analysis.
Our approach encompasses the introduction of novel methodologies and extensive ablation studies. These studies critically assess various aspects such as loss functions, box regression techniques, and classification strategies to refine the model's precision in object detection. The paper details the experimental application of the LSKNet backbone in synergy with the DiffusionDet heads, a combination tailored to meet the specific challenges in aerial image object detection.
The findings of this research indicate a substantial enhancement in the model's performance, especially in the accuracy-time tradeoff. The proposed model achieves a mean average precision (MAP) of approximately 45.7\%, which is a significant improvement, outperforming the RCNN model by 4.7\% on the same dataset. This advancement underscores the effectiveness of the proposed modifications and sets a new benchmark in aerial image analysis, paving the way for more accurate and efficient object detection methodologies. The code is publicly available at \url{https://github.com/SashaMatsun/LSKDiffDet}
\end{abstract}

\section{Introduction}
\label{sec:intro}

Object detection in aerial imaging has emerged as a dynamic and pivotal area of research, focusing on identifying and localizing objects within high-resolution images captured via airborne platforms, such as satellites, drones, or aircraft \cite{cheng2016survey}. This technology finds application in a diverse range of fields, including but not limited to urban planning \cite{zhang2016building}, precision agriculture \cite{mali2017high}, disaster management \cite{cheng2018automatic}, and military surveillance \cite{li2018deep}. The integration of cutting-edge machine learning methodologies, notably deep learning and convolutional neural networks \cite{girshick2014rich}, enables these object detection models to process extensive aerial datasets efficiently, identifying specific objects like vehicles, buildings, and vegetation \cite{li2017vehicle}. However, this domain faces several challenges, such as dealing with varying image resolutions \cite{zhou2016cad}, managing occlusions \cite{han2018optimized}, the necessity for large and accurately annotated training datasets \cite{lin2014microsoft}, and the real-time processing of high-resolution images \cite{redmon2016you}. Addressing these challenges is essential for fully unlocking the potential of aerial imaging in object detection, thereby facilitating more effective, data-informed decision-making across a variety of sectors \cite{cheng2016survey}.
\par
In this paper, we present a series of innovative contributions that significantly propel the field of aerial image analysis forward. Our comprehensive and multifaceted approach includes the introduction of a new backbone architecture, the incorporation of diffusion models, and the application of a variety of loss functions. In addition, we explore the impact of different activation functions and the refinement of hyperparameters to achieve optimal performance. These elements collectively represent a leap in advancing aerial image analysis, as detailed below:

\begin{enumerate}[noitemsep,nolistsep,partopsep=5pt,topsep=5pt]
\item We employ \textit{large kernel convolutions} and \textit{spatial kernel selection} in tandem with a \textit{feature pyramid network} (FPN)~\cite{lsknet}, constructing a robust and effective backbone for aerial image analysis. This novel design significantly enhances feature extraction and representation in aerial imagery.

\item The adaptation of the \textit{diffusion model} (DiffusionDet) \cite{Diffusiondet} to aerial imaging, with tailored modifications, leads to marked improvements in object detection accuracy within complex aerial scenes.

\item We introduce an innovative and refined model architecture that substantially elevates the accuracy of aerial image analysis. These modifications result in a more powerful and efficient model tailored to aerial imaging contexts.

\item Extensive experimentation is conducted to evaluate the impact of various \textit{activation functions} on our model's performance. This investigation identifies the most effective activation function, boosting the model's overall performance and robustness.

\item To tackle the prevalent class imbalance issue, we develop a \textit{weighted focal loss} function and examine the adaptation of other loss functions for box regression. This approach effectively addresses class imbalances, simultaneously enhancing the model's accuracy.

\item We thoroughly analyse the influence of \textit{hyperparameters} and \textit{post-processing methods}, fine-tuning them to optimize results. This meticulous optimization process ensures our proposed model's peak performance in the complex aerial image analysis domain.

\end{enumerate}

Our work introduces an expansive and innovative methodology for aerial image analysis, achieving superior performance while surpassing existing methods. A key aspect of our approach is its efficiency in resource utilization. Despite using a limited number of GPUs and fewer iterations, our model demonstrates significant advancements in accuracy and robustness. This efficient use of computational resources underscores our model's effectiveness and highlights its sustainability, making it a more energy-efficient solution in aerial image analysis. By meticulously addressing various facets of model design and optimization, our study sets a new benchmark in delivering high-performance aerial image analysis with a reduced environmental impact.

\section{Dataset}
In this work, we used a patchified version of the Instance Segmentation in Aerial Images Dataset (iSAID)\cite{isaid}, which is a comprehensive dataset designed specifically for aerial image object detection and instance segmentation tasks. It contains diverse high-resolution aerial images collected from various sources, including satellites and unmanned aerial vehicles (UAVs). The dataset encompasses a wide range of scenarios, such as urban, rural, and natural environments, providing a robust foundation for training and evaluating object detection and segmentation models.
\par
The original version of iSAID contains 655,451 object instances belonging to 15 categories within 2,806 high-resolution images. The images within the iSAID dataset are the same as those in the DOTA-v1.0 dataset~\cite{xia2018dota}, which are primarily collected from Google Earth. Some of the images are captured by the JL-1 satellite, while others are taken by the GF-2 satellite, both operated by the China Centre for Resources Satellite Data and Application.
\par
Our study employed a patchified version of the iSAID dataset, which includes 28029 images. These were derived by segmenting the original dataset into patches of size $800 \times 800$. This segmentation led to a substantial increase in the number of instances in the training subset, amounting to 704428. The rise in instance count can be largely attributed to overlaps in the dataset patches. Using this batchified approach provided two key advantages: first, it significantly enriched the training data's diversity and complexity, thereby enhancing our model's robustness in various aerial imaging scenarios. Second, it allowed for a more comprehensive training experience, exposing the model to various instances and contextual environments. It is crucial for improving its generalization capabilities across diverse aerial images. Crucially, despite the segmentation and augmentation process, we meticulously maintained the separation of the train, validate, and test sets, ensuring no data leakage between them. This careful partitioning guarantees the integrity and reliability of our evaluation process.
\par
\begin{figure}[]
\centering
\begin{tabular}{@{}c@{\hspace{1mm}}c@{}}
 \includegraphics[width=0.48\linewidth]{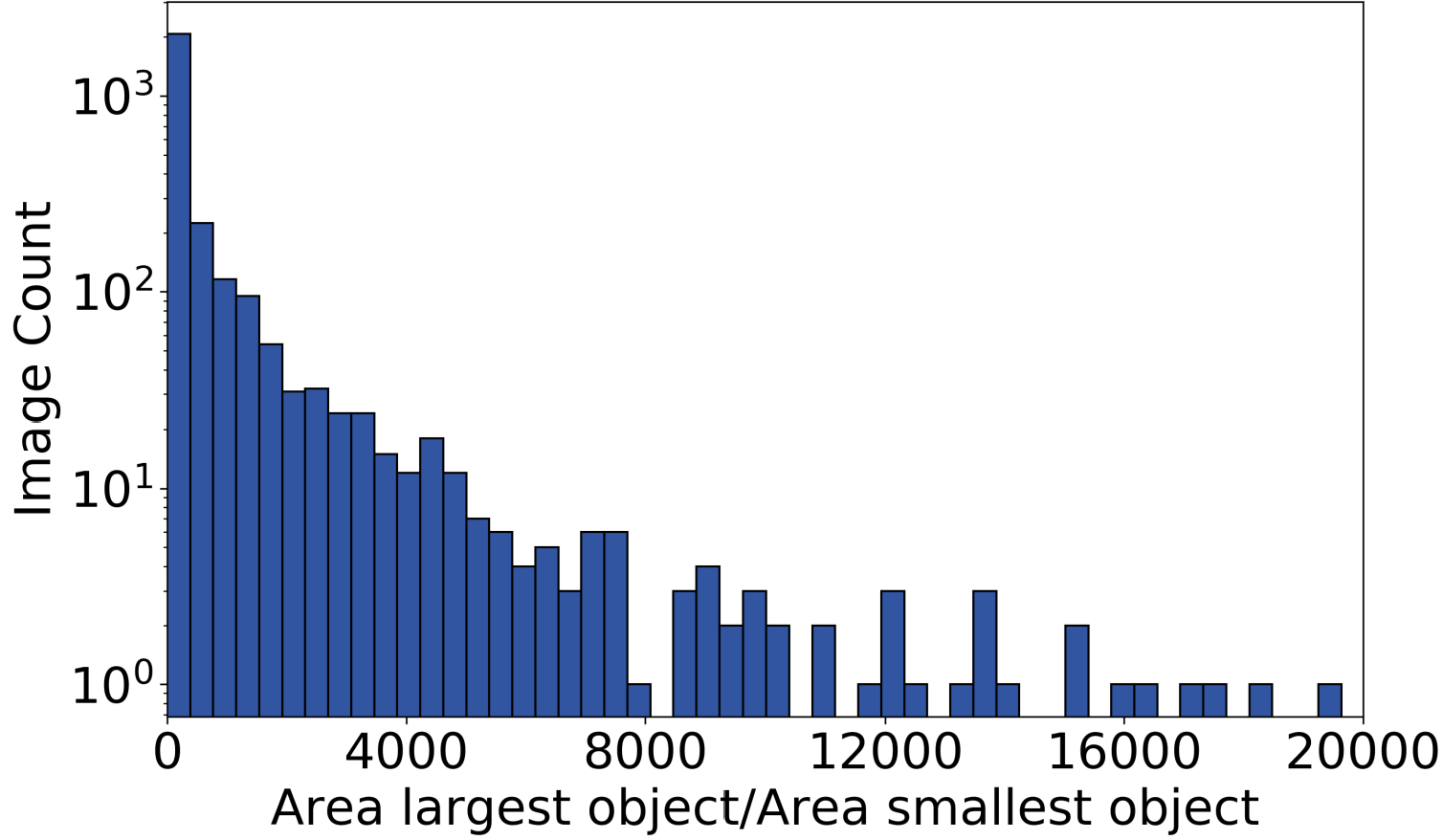} &
 \includegraphics[width=0.48\linewidth]{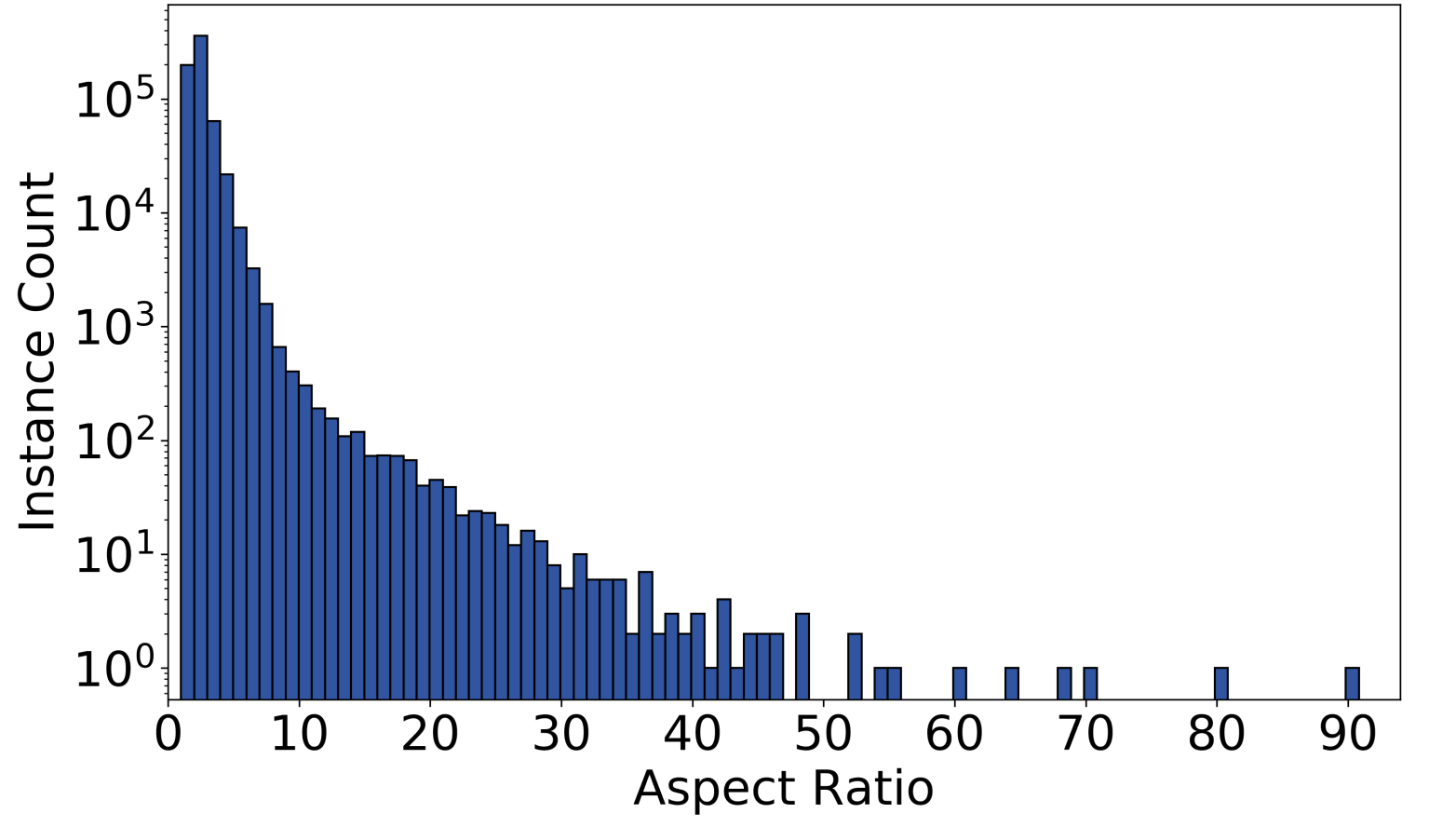} \\
(a) & (b)
\end{tabular}
\caption{(a) Ratio between the largest and smallest objects' areas shows scale variation. (b) Aspect ratio variations between instances.} 
\label{variation}
\end{figure}

Despite its richness, the iSAID dataset presents several challenges to researchers and practitioners in the field of aerial object detection. Addressing these challenges requires the development of more advanced and robust machine-learning models capable of handling the unique complexities of aerial imagery. These challenges include:

\begin{enumerate}[noitemsep,nolistsep,partopsep=5pt,topsep=5pt]
    \item \textit{Varied Object Sizes:} Aerial images feature objects of diverse sizes, from large buildings to small vehicles, making it difficult for models to accurately identify and segment instances.
    \item \textit{Occlusion:} Objects in aerial images often overlap or are partially hidden by other objects or natural features, making it challenging to segment and detect them accurately.
    \item \textit{Scale Variation:} The dataset contains images with varying scales and aspect ratios, which can impact the performance of models trained on this data, as shown in Figure~\ref{variation}.
    \item \textit{Complex Backgrounds:} Aerial images often feature intricate and cluttered backgrounds, making it difficult for models to distinguish between objects and their surroundings.
    \item \textit{Illumination and Weather Conditions:} Changes in lighting and atmospheric conditions can impact the visibility and appearance of objects in aerial images, posing additional challenges for object detection and segmentation models.
\end{enumerate}

\section{Related Works}

Aerial imaging and object detection have evolved into indispensable tools across various applications, offering profound insights into terrestrial phenomena and human activities. The advent of remote sensing technologies, encompassing satellite and drone imagery, has revolutionized observational and analytical methods in large-scale environmental and geographical studies \cite{yang2018object}. Aerial imaging facilitates the acquisition of high-resolution data over extensive areas, proving invaluable in diverse tasks like land use classification \cite{zhong2017learning}, disaster response \cite{cheng2018multi}, and environmental monitoring \cite{singh2017urban}. In this context, object detection in aerial images is pivotal for distilling actionable information from these vast datasets, aiding in identifying and localizing various objects and features.
\par
Recent advancements in deep learning-based object detection algorithms have markedly enhanced the accuracy and efficiency of analyzing aerial images \cite{xia2018dota, lamm2018xview}. The synergy of aerial imaging and object detection has profoundly influenced numerous fields, contributing significantly to our understanding and capabilities in addressing global challenges.
\par
One-stage object detection models have gained prominence due to their operational efficiency and efficacy. These models amalgamate object localization and classification tasks into a single streamlined network, thereby reducing inference time. A notable example of this model type is the You Only Look Once (YOLO) framework \cite{redmon2016you}, which segments the input image into a grid system, predicting bounding boxes and class probabilities for each grid cell. Similarly, the Single Shot MultiBox Detector (SSD), proposed by Liu et al. \cite{liu2016ssd}, employs multi-scale feature maps to handle objects of various sizes effectively. These one-stage models have shown commendable performance in a range of real-world scenarios.
\par
Conversely, two-stage object detection models typically consist of a region proposal network followed by a classification network. The initial stage generates candidate regions for objects, while the subsequent stage classifies these regions into specific object categories. The Region-based Convolutional Networks (R-CNN) family \cite{girshick2014rich} exemplifies two-stage models. The R-CNN model utilizes a selective search for region proposal generation, followed by classification via a CNN. Successors like Fast R-CNN \cite{girshick2015fast} and Faster R-CNN \cite{ren2015faster} have enhanced the efficiency and accuracy of the original R-CNN through innovations like ROI pooling and region proposal networks.
\par
State-of-the-art in object detection within aerial imagery has progressed rapidly, spurred by the availability of large-scale aerial image datasets like DOTA \cite{xia2018dota} and xView \cite{lamm2018xview}. These datasets have enabled focused training and evaluation of object detection models tailored for aerial contexts. Aerial images often pose unique challenges, including significant object scale variations, cluttered backgrounds, and diverse viewpoints. To address these challenges, specialized models have been developed, such as the Oriented Region Proposal Network (ORPN) \cite{zhang2016arbitrary}, designed to detect objects irrespective of their orientations. The High-Resolution Network (HRNet) \cite{wang2019deep} effectively manages objects of varying sizes by maintaining high-resolution feature maps throughout the model. Furthermore, the Adaptively-Sized Object Detector (ASOD) \cite{wang2020asod} adapts to varying object sizes by adjusting receptive fields and anchor scales. These advancements have substantially improved object detection in aerial images, enhancing the accuracy and efficiency of data analysis.

\section{Methodology}
In this section, we elucidate the technical intricacies of the methods employed in constructing our model.

\subsection{Model Architecture}
\subsubsection{LSKNET}
The backbone of our model mirrors the general architecture of prevalent models such as those discussed in \cite{liu2022convnet}, consisting of repeated blocks with a similar structure. A key innovation in our approach is integrating the Large Selective Kernel (LSK) mechanism \cite{li2023large} into each backbone block. This integration is pivotal in enhancing the feature extraction capability of the model by providing a broader contextual area.

\textit{Large Kernel Convolutions}: Implemented as a sequence of depthwise convolution layers \cite{chollet2017xception}, these convolutions utilize increasing kernel sizes and dilation rates. This configuration rapidly expands the receptive field, as detailed in \cite{ding2022scaling, liu2022more}. The structure's primary benefits are twofold: it facilitates extracting multiple features encompassing various contextual areas. It offers superior efficiency to a single large kernel with an equivalent receptive field. For instance, for an input with 64 channels, a sequential mechanism with a structure of \((3,1) \rightarrow (5,2) \rightarrow (7,3)\) requires only 11.3K parameters. In contrast, a single convolution layer of size 29 would necessitate 60.4K parameters.

\textit{Spatial Kernel Selection}: As per \cite{li2019selective}, this process dynamically selects kernels suited to different objects based on the extracted features. Initially, features from various-sized kernels are concatenated into a feature map of size \(\mathbb{R}^{h \times w \times N}\). Subsequently, the channel-wise average and maximum are computed and integrated into a feature descriptor of size \(\mathbb{R}^{h \times w \times 2}\). A convolution layer followed by a sigmoid activation function is then applied to these feature descriptors, resulting in a spatial attention map of size \(\mathbb{R}^{h \times w \times N}\). The final output of this module is the element-wise product of the concatenated input feature maps and the spatial attention map.

\begin{figure}[]
\centering
\caption{(a) The original LSK Module (b) Our modified LSK Module with a residual connection.} 
\begin{tabular}{cc}
 \includegraphics[width=0.42\linewidth]{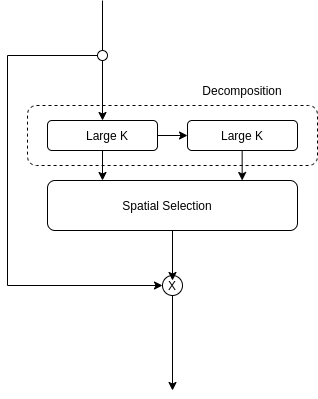}  &  
 \includegraphics[width=0.48\linewidth]{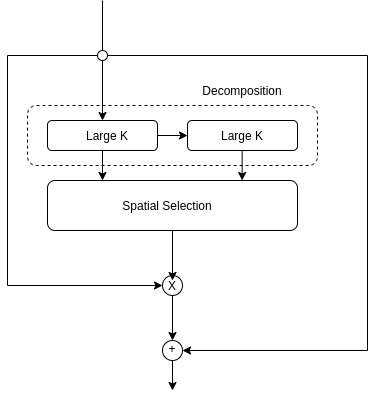} \\
(a)   &  (b) 
\end{tabular}
\label{lsk}
\end{figure}

Finally, \textit{a feature pyramid network} \cite{lin2017feature} is constructed using a series of downsampling blocks, each comprising a sequence of Large Selective Kernel blocks. This configuration ensures that the final output of the backbone comprises multiple feature maps of varying resolutions obtained by passing the input through a differing number of blocks. As an innovative modification, we have also incorporated residual connections parallel to the spatial filtering operation. This addition allows the preservation and passage of features potentially filtered out by the preceding LSK block. Figure \ref{lsk} shows our modification to the LSK block compared to the original block.

\subsubsection{DiffusionDet}
We chose DiffusionDet head for our model~\cite{Diffusiondet}. DiffusionDet is a novel framework that approaches object detection as a denoising diffusion process, transitioning from noisy boxes to actual object boxes. During the training phase, object boxes diffuse from ground-truth boxes to a random distribution, and the model learns to reverse this process. During inference, the model progressively refines a set of randomly generated boxes to produce the final output. Comprehensive evaluations on standard benchmarks, including MS-COCO and LVIS, demonstrate that DiffusionDet outperforms many well-established detectors. This work reveals two key insights in object detection: First, random boxes, despite being significantly different from predefined anchors or learned queries, can still serve as effective object candidates. Second, object detection, as a representative perception task, can be addressed using a generative approach.

\par
The diffusion model generates data samples iteratively, requiring multiple runs of the model $f_{\theta}$ during the inference stage. To address the computational intractability of applying $f_{\theta}$ on the raw image at every step, the model is divided into an image encoder and a detection decoder. The image encoder extracts high-level features from the raw input image, while the detection decoder refines box predictions from noisy boxes using these features. Inspired by Sparse R-CNN~\cite{Sparse}, the detection decoder takes in proposal boxes, crops RoI-features from the feature map, and sends them to the detection head for box regression and classification. The main differences between DiffusionDet and Sparse R-CNN include the use of random boxes instead of learned ones, the input requirements, and the re-use of the detector head in iterative sampling steps with shared parameters across different steps.
\par
We prefer this model over others in this coherence because of its ability to handle noisy images and pay attention to small objects. Those are two main challenges in aerial images, so we think it may be suitable for that task. We replaced the default Swin transformer backbone with our modified version of LSKNet and initialized the model using pre-trained weights from COCO dataset.

\subsection{Augmentations}
Data augmentation stands as a cornerstone technique in enhancing machine learning models' performance, particularly in aerial image analysis. This process involves generating new training samples by applying a range of transformations to existing images. Such transformations, which include rotation, scaling, flipping, and colour modifications, substantially increase the training dataset's diversity. This, in turn, equips models with better generalization capabilities for new, unseen data.

Our study focused on two prominent data augmentation techniques: flipping and Albumentations. Flipping, a simple yet effective method, creates new training instances by mirroring the original images horizontally or vertically. This approach aids in diversifying the dataset and plays a crucial role in mitigating the risk of overfitting.

Albumentations, a specialized data augmentation library tailored for computer vision tasks \cite{albumentations}, was another key component of our methodology. This library offers a comprehensive suite of image transformations designed to enhance the model's generalization ability. These transformations encompass geometric operations, such as rotations, translations, scaling, flipping, and photometric adjustments, like altering brightness, contrast, and colour. 

Applying these data augmentation techniques is particularly beneficial in addressing the unique challenges posed by aerial images. These challenges include variations in scale and resolution, geometric distortions, diverse environmental conditions, and seasonal changes. By employing flipping and Albumentations, our machine learning models are better equipped to adapt to and accurately interpret the complex characteristics of aerial imagery.

\subsection{Loss Functions}
Loss function selection is a pivotal aspect of object detection, fundamentally guiding the learning process. Broadly, loss functions in object detection can be categorized into two types: bounding box regression loss and classification loss. The bounding box regression loss measures the similarity between predicted and ground truth bounding boxes, considering attributes such as shape, orientation, aspect ratio, and centre distance. Various loss functions, or their linear combinations, are utilized for this purpose:

\begin{enumerate}
    \item \textit{Intersection Over Union (IOU)}: A widely-used metric for evaluating the accuracy of object detection models. The IOU loss is defined as:
    \begin{equation}
        IOU_{loss} = 1 - \frac{A_{\text{pred}} \cap A_{\text{gt}}}{A_{\text{pred}} \cup A_{\text{gt}}}
    \end{equation}
    
    \item \textit{Generalized Intersection Over Union (GIOU)}: An extension of IOU, GIOU considers the smallest convex hull that encloses both the ground truth and predicted boxes. It is more robust than IOU, as it accounts for the shape and orientation of the boxes, thereby reducing the impact of misaligned boxes on the final loss value~\cite{GIOU}.
    \begin{equation}
        GIOU_{loss} = 1 - (IOU - \frac{A_{C} - A_{U}}{A_{C}})
    \end{equation}
    
    \item \textit{Complete Intersection Over Union (CIOU)}: CIOU further enhances GIOU by incorporating the aspect ratio and centre distance between the ground truth and predicted boxes. This integration improves convergence and localization accuracy, particularly beneficial for objects with varying aspect ratios~\cite{CIOU}.
    \begin{equation}
        CIOU = 1 - (GIOU - \alpha \cdot \frac{d^2_{\text{center}}}{A_{C}} - \beta \cdot \frac{v^2_{\text{aspect\_ratio}}}{1 - IOU})
    \end{equation}

    \item \textit{SmoothL1 Loss}: A variant of L1 loss, SmoothL1 Loss is less sensitive to outliers. It applies a smooth approximation to the absolute function, transitioning from L1 to L2 loss near the origin. This approach results in a more stable learning process and mitigates the impact of noisy samples.
    \begin{equation}
        L_{\text{SmoothL1}}(x) = \begin{cases} 0.5x^2 & \text{if } |x| < 1 \\ |x| - 0.5 & \text{otherwise} \end{cases}
    \end{equation}
\end{enumerate}

Various loss functions are applicable for classification tasks, with Focal Loss being of particular interest. Focal Loss is designed to address the class imbalance in object detection tasks by introducing a modulating factor that down-weights the contribution of easy examples and concentrates on harder, misclassified examples~\cite{focal}.
\begin{equation}
    FL(p_t) = -\alpha_t (1-p_t)^{\gamma} \log(p_t)
\end{equation}

\begin{figure}[h]
\centering
 \includegraphics[width=0.9\linewidth]{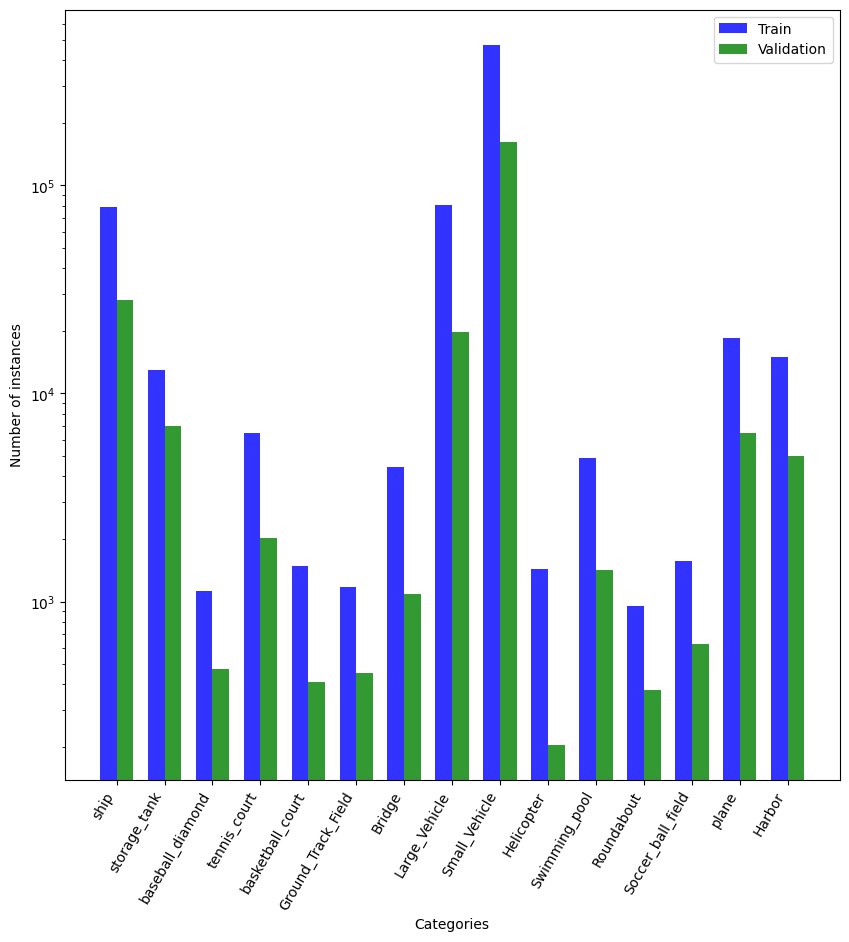}  
\caption{number of instances in each class for train and validation} 
\label{classes}
\end{figure}

As Figure~\ref{classes} illustrates, significant class imbalance might limit Focal Loss's efficacy. To address this, we implement Weighted Focal Loss, combining the principles of Focal Loss with class weighting. This method assigns varying weights to each class, enabling the model to prioritize minor classes or those with higher misclassification costs. This approach can enhance overall performance, especially in pronounced class imbalances. Equation~\ref{weighted} depicts the Weighted Focal Loss, where $\alpha_t$ is the weighting factor for the target class, calculated for simplicity as the inverse ratio of the number of samples of each class to the total number of samples. 
\begin{equation}
    WFL(p_t) = -\alpha_t w_c (1-p_t)^{\gamma} \log(p_t)
    \label{weighted}
\end{equation}

\subsection{Activation Function}
Activation functions are crucial in neural networks, introducing non-linearity that enables models to learn complex patterns from input data. This paper discusses three advanced activation functions: Mish, Hardswish, and Gaussian Error Linear Units (GELU), each bringing unique benefits to the model's learning capability.

\begin{enumerate}
    \item \textit{Mish}: Mish is a self-regularized novel activation function that has demonstrated superior performance to traditional functions like ReLU, Leaky ReLU, and Swish \cite{mish}. It introduces attributes of smoothness and non-monotonicity, fostering enhanced gradient flow and expedited convergence. The Mish function is defined as:
    \begin{equation}
        \text{Mish}(x) = x \cdot \tanh(\text{softplus}(x))
    \end{equation}
    This function has effectively promoted deeper feature extraction with improved learning dynamics.

    \item \textit{Hardswish}: As a computationally efficient alternative to the Swish activation function, Hardswish offers comparable performance while reducing computational overhead \cite{hardswish}. It has found utility in lightweight models such as MobileNetV3 and EfficientNet, which aim to balance high accuracy with low computational complexity. The following equation defines Hardswish:
    \begin{equation}
        \text{Hardswish}(x) = x \cdot \frac{\text{ReLU6}(x + 3)}{6}
    \end{equation}
    The primary advantage of Hardswish lies in its efficiency, particularly in resource-constrained environments.

    \item \textit{Gaussian Error Linear Unit (GELU)}: Inspired by the Gaussian error function, GELU has gained popularity in various models, including BERT and GPT, especially in natural language processing tasks \cite{gelu}. It is characterized by:
    \begin{equation}
        \text{GELU}(x) = x \cdot \frac{1}{2} \left( 1 + \text{erf} \left( \frac{x}{\sqrt{2}} \right) \right)
    \end{equation}
    GELU is renowned for its ability to facilitate more nuanced and probabilistic feature transformations, contributing to the model's overall expressiveness and performance.
\end{enumerate}

\subsection{Hyper-parameters}

There are many hyper-parameters that can be tuned to increase performance. From these, we made an ablation study on the effect of each, which are:
\begin{enumerate}
    \item \textit{The learning rate} determines the step size taken during gradient descent optimization. A crucial hyperparameter influences the convergence rate and model performance~\cite{lr}. A too-large learning rate can lead to divergence, while a too-small learning rate can result in slow convergence.
    \item \textit{Number of proposals} refers to the number of candidate bounding boxes generated by a region proposal network (RPN) in object detection models\cite{ren2015faster}. This hyperparameter affects the trade-off between recall and computational complexity.
    \item \textit{Aspect ratios} are the different proportions of anchor boxes used in object detection models~\cite{redmon2016you}. They help the model detect objects with varying shapes and sizes.
    \item \textit{The number of epochs} is the number of times the entire training dataset is processed during training. A higher number of epochs can result in better model performance, but it may also increase the risk of overfitting if the model is trained for too long. Choosing the optimal number of epochs depends on the specific problem and dataset.
    \item \textit{Batch size} is the number of training samples used to compute the gradient during a single optimization step~\cite{batch}. Larger batch sizes can lead to more stable gradient estimates but may require more memory and computational resources.
    \item \textit{Images per batch} is the number of images used in each batch during training. This hyperparameter is related to the batch size and influences the memory requirements and the stability of the gradient estimates.

\end{enumerate}

\setcounter{table}{4}
\begin{table*}
    \caption{Results of Best Model on Validation \& Test Sets: Notations: ST: Storage tank, TC: Tennis court, BD: Baseball Diamond, BC: Basketball court, GTF: Ground Track Field, LV: Large Vehicle, SV: Small Vehicle, HC: Helicopter, SP: Swimming pool, RA: Roundabout, SBF: Soccer ball field. }
\setlength{\tabcolsep}{4pt} 
  {\fontsize{8}{11}\selectfont 
\label{results}
    \centering
    \begin{tabular}{ | c | c | c | c | c | c | c | c | c | c |c | c |c | c | c |  c |  c |  c |  c |   c |  }
      \hline
      \thead{Data}  & \thead{AP} & \thead{AP75} & \thead{AP50}   & \thead{Ship}  & \thead{ST}  &  \thead{TC} & \thead{BD}  & \thead{BC} & \thead{GTF} &  \thead{Bridge} & \thead{LV} & \thead{SV} & \thead{HC}  & \thead{SP}  & \thead{RA} & \thead{SBF}  & \thead{Plane}  & \thead{Harbor}  \\
            \hline
      \makecell{Test}& \makecell{45.7}&  \makecell{50.6}  & \makecell{66.8}   &\makecell{55.4}  &\makecell{33.4}  &\makecell{50.9}  &\makecell{25.0}  &\makecell{66.1}  &\makecell{34.1}  &\makecell{45.3}  &\makecell{58.5}  &\makecell{37.2}  &\makecell{46.6}  &\makecell{22.4}  &\makecell{25.0}  &\makecell{45.3}  &\makecell{78.8}  &\makecell{45.7}   \\ 
        \hline

      \makecell{Validation}& \makecell{44.8}&  \makecell{48.6}  & \makecell{67.0}  &\makecell{57.8}   &\makecell{37.6}  &\makecell{80.9}  &\makecell{55.1}  
      &\makecell{ 44.5}  &\makecell{49.1}  &\makecell{25.2}  &\makecell{47.3}  &\makecell{20.8}  
      &\makecell{21.3}  &\makecell{38.2}  &\makecell{34.3}  &\makecell{49.6}  &\makecell{70.8}  &\makecell{49.8}   \\ 
    \hline
              
    \end{tabular}\label{best}}
  \end{table*}
  
\section{Experiment Setup}
To comprehensively understand the impact of each modification in our model, we adopted a systematic approach in our experiments, altering only one variable at a time. This systematic process allowed us to isolate and examine the specific effect of each change.

Our initial focus was on the model architecture. We explored the combination of LSKNet as the backbone with DiffusionDet heads. We conducted five distinct experiments to assess the impact of varying the backbone and the head. In each, we altered either the backbone or the head only. Specifically, we utilized ResNet and LSKNet as backbones with the Faster RCNN and replicated this configuration for DiffusionDet. Additionally, Swin Transformer was employed as the backbone in conjunction with DiffusionDet in our fifth experiment.

For all subsequent experiments, we standardized the architecture, employing LSKNet as the backbone and DiffusionDet for the heads. This model configuration included GeLU as the activation function and utilized focal loss and GIOU as loss functions. By default, the model also employed Non-Maximum Suppression (NMS) and default aspect ratios of [0.5, 2, 4].

The hyper-parameters were set as follows: batch size = 512, number of images per batch = 3, learning rate = 0.00005, number of proposals = 300, and a maximum of 100000 iterations. Given our limited hardware resources, specifically the constraint of utilizing only a single GPU with 24 GB of memory, we were mindful of the complexities added to our model. This hardware limitation influenced decisions regarding the number of images per batch, iterations, and other aspects that could potentially increase performance. Each modification was underpinned by a rationale aimed at performance improvement:
\begin{itemize}
    \item \textit{Activation Functions:} Mish was selected to address the vanishing gradient problem, known to enhance model accuracy, especially in deep networks. Hard Swish, recognized for its computational efficiency, has been shown to yield similar or slightly better accuracy than ReLU in certain tasks.
    \item \textit{Architectural Modifications:} We expanded the model by adding a block with a depth of 32 at the start of the model's sequence. This expansion aimed to enhance the detection of smaller objects, leveraging high-resolution feature maps. Furthermore, to prevent the potential loss of important features by spatial selection, we experimented with direct residual connections from the unfiltered feature map to the output of the LSK block.
    \item \textit{Loss Functions:} CIOU was chosen for considering overlaps, aspect ratios, and centre distances between predicted and ground-truth boxes. Weighted Focal Loss targeted class imbalance by focusing on harder examples. Smooth L1 Loss was introduced to mitigate the effect of outliers, potentially improving regression performance.
\end{itemize}

Additional empirical adjustments to the hyper-parameters were made based on specific characteristics of the dataset and expected outcomes:
\begin{itemize}
    \item \textit{Aspect Ratios:} Adjusted to [0.25, 0.75, 2, 4] to better suit the dataset's large variance in aspect ratios.
    \item \textit{More Proposals:} Increased to 700, aiming to enhance accuracy by offering a broader range of regions for evaluation.
    \item \textit{More Images Per Batch:} Increased from 3 to 4 to provide a richer variety of data for the algorithm, enhancing learning and generalization.
    \item \textit{Soft NMS:} Implemented as an alternative to traditional NMS, this technique lowers scores of overlapping boxes rather than discarding them, potentially retaining more accurate predictions in scenarios with closely packed or partially occluded objects.
\end{itemize}


\section{Results \& Discussion}

\begin{figure*}[]
\centering
\begin{tabular}{@{}c@{\hspace{1mm}}c@{\hspace{1mm}}c@{\hspace{1mm}}c@{}}
 \adjustbox{margin=0pt}{\includegraphics[width=0.22\linewidth]{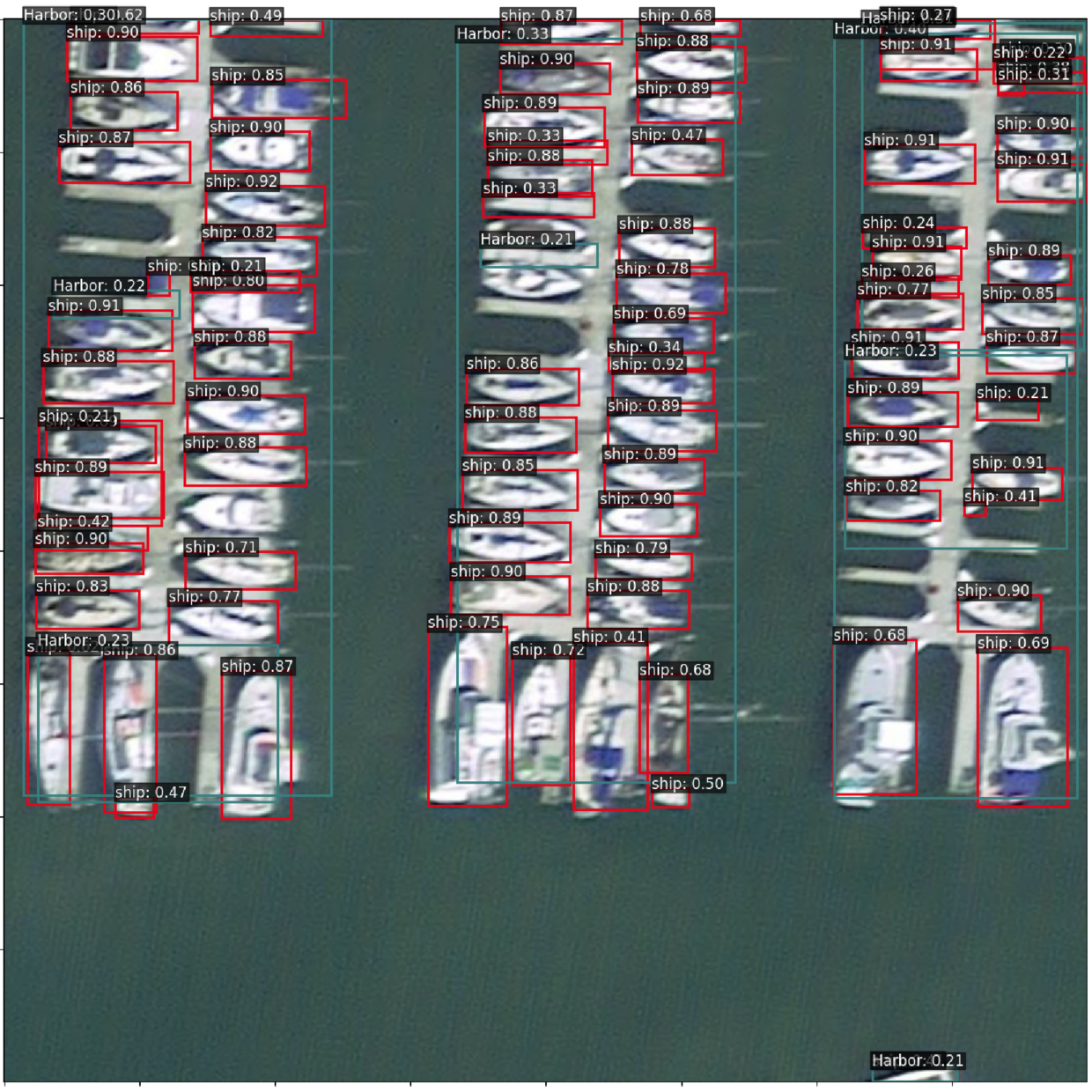}} &
 \adjustbox{margin=0pt}{\includegraphics[width=0.22\linewidth]{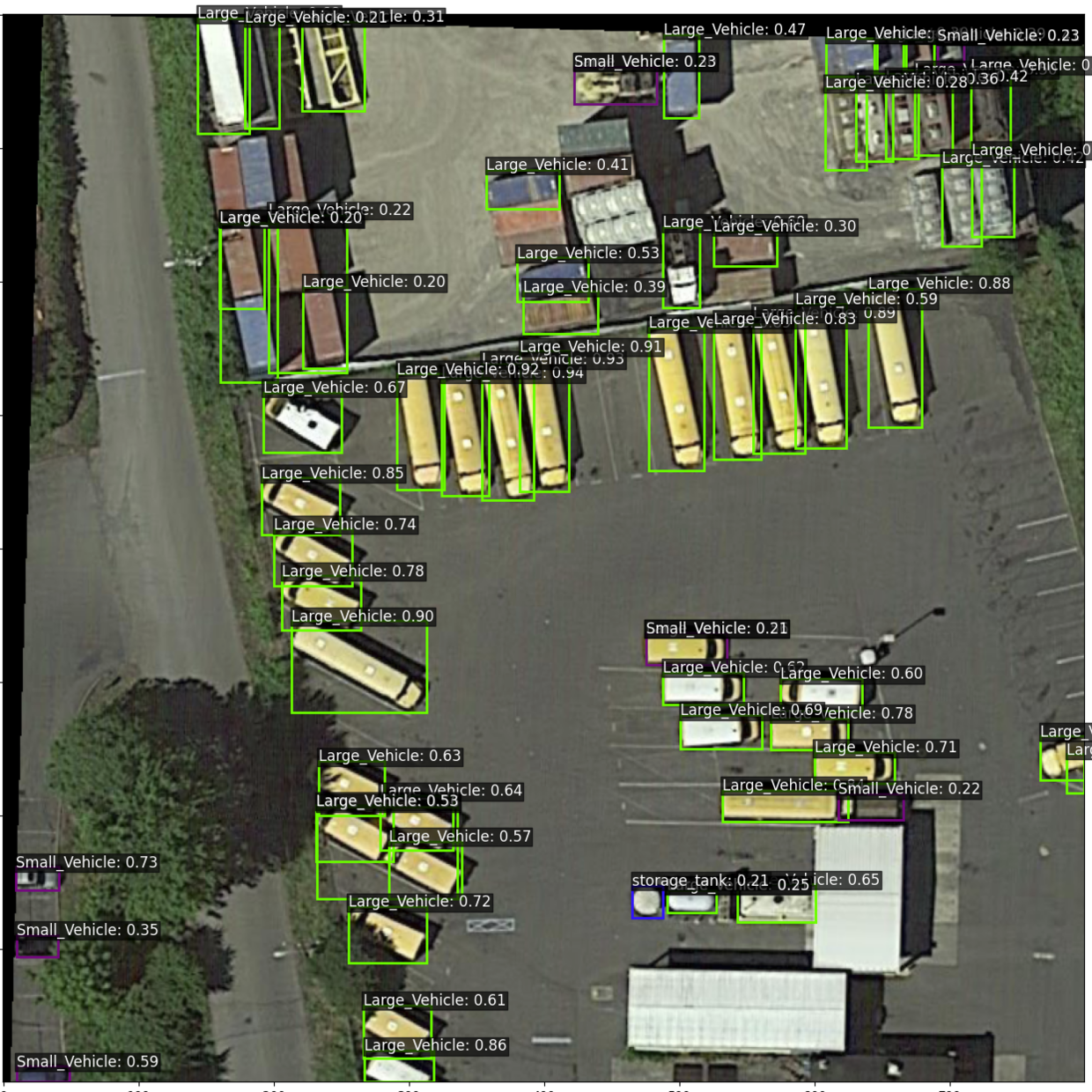}} &
 \adjustbox{margin=0pt}{\includegraphics[width=0.22\linewidth]{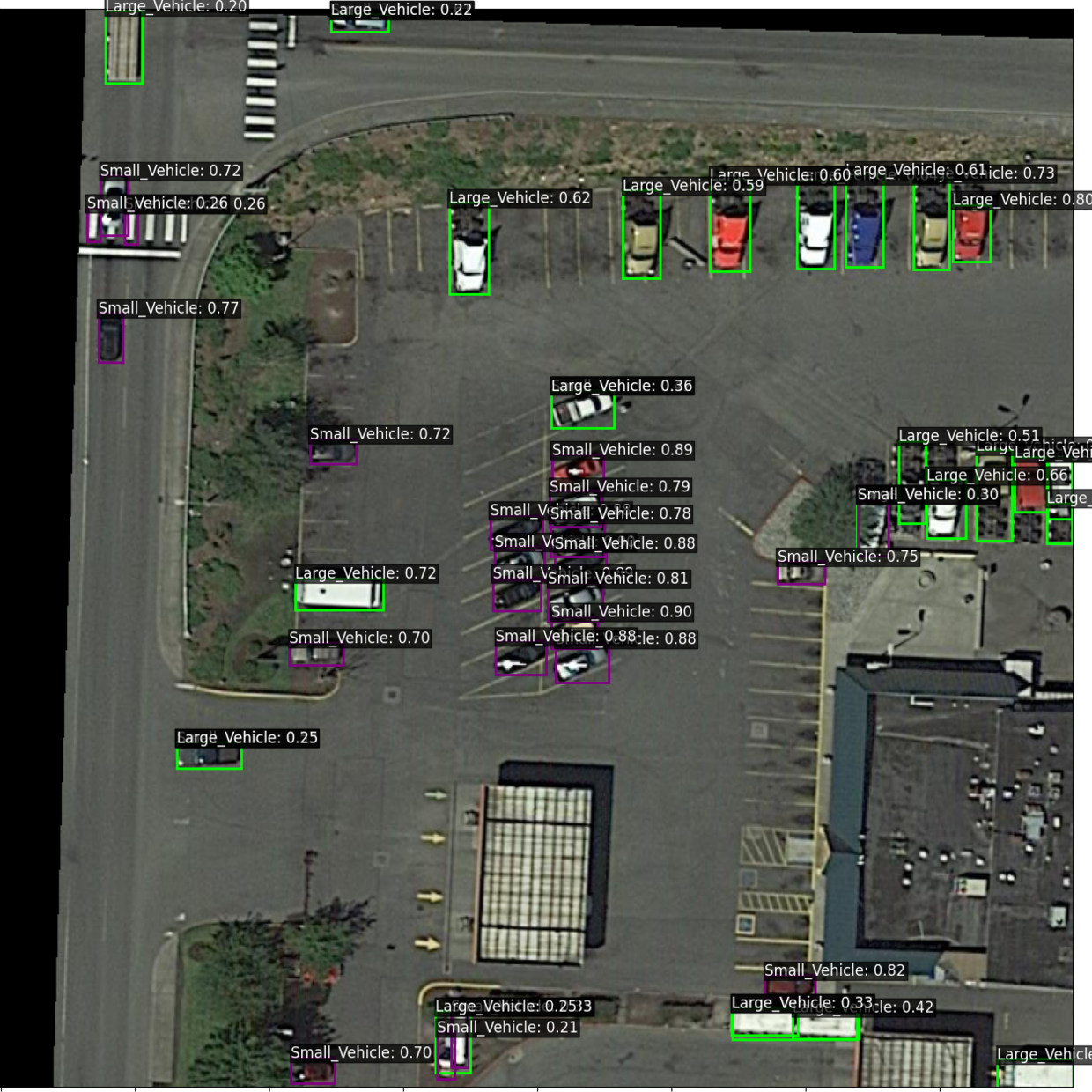}} &
 \adjustbox{margin=0pt}{\includegraphics[width=0.22\linewidth]{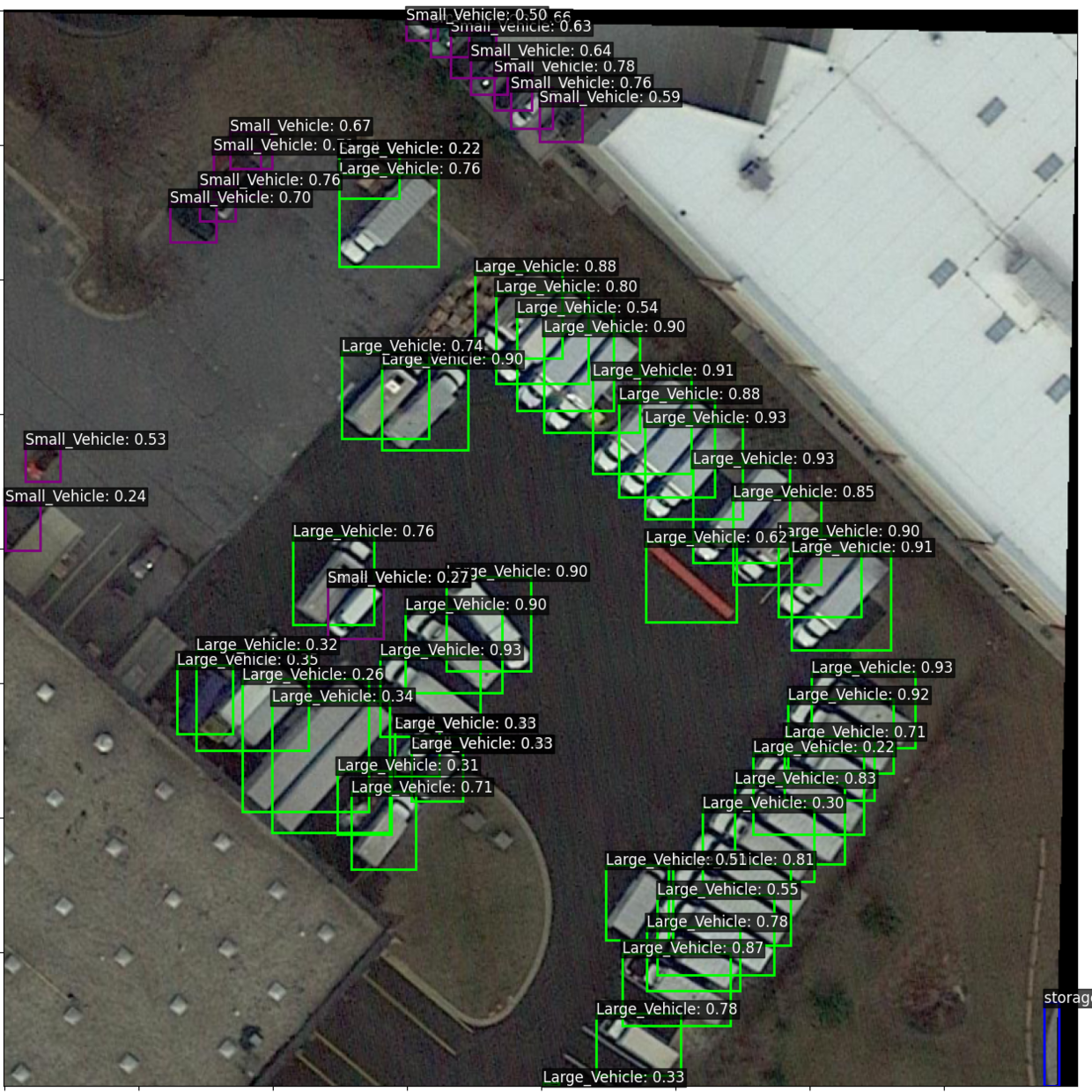}} \\
 \adjustbox{margin=0pt}{\includegraphics[width=0.22\linewidth]{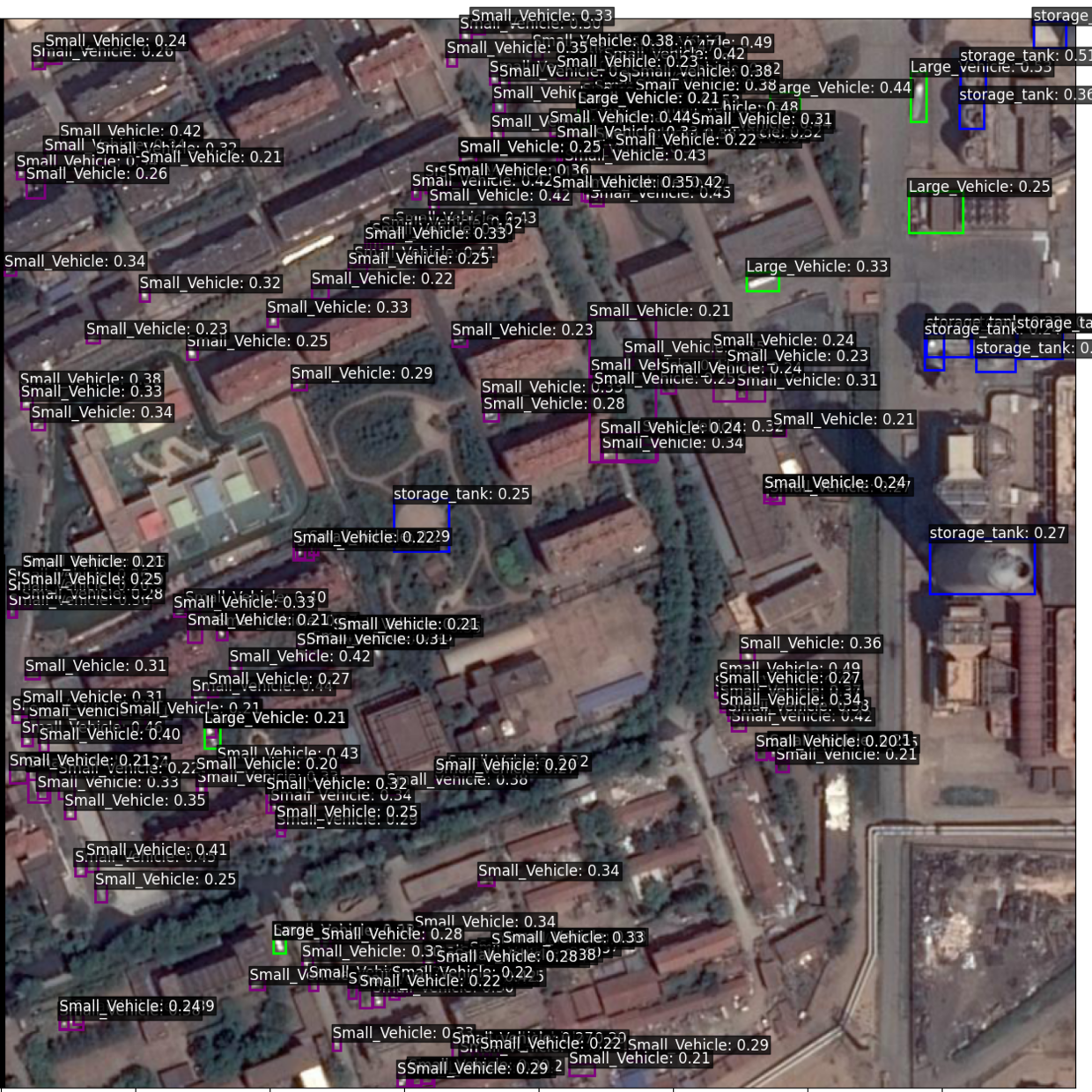}} &
 \adjustbox{margin=0pt}{\includegraphics[width=0.22\linewidth]{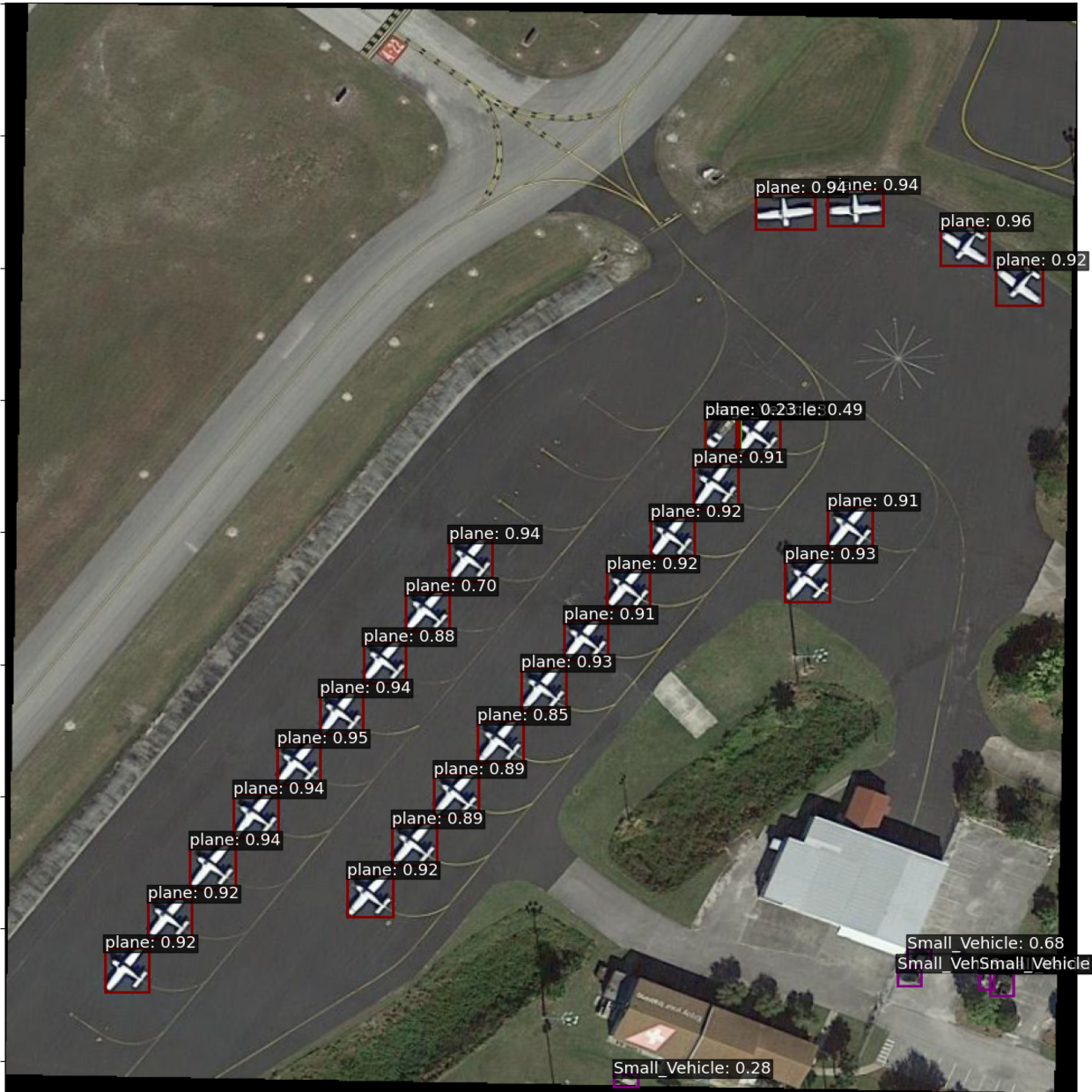}} &
 \adjustbox{margin=0pt}{\includegraphics[width=0.22\linewidth]{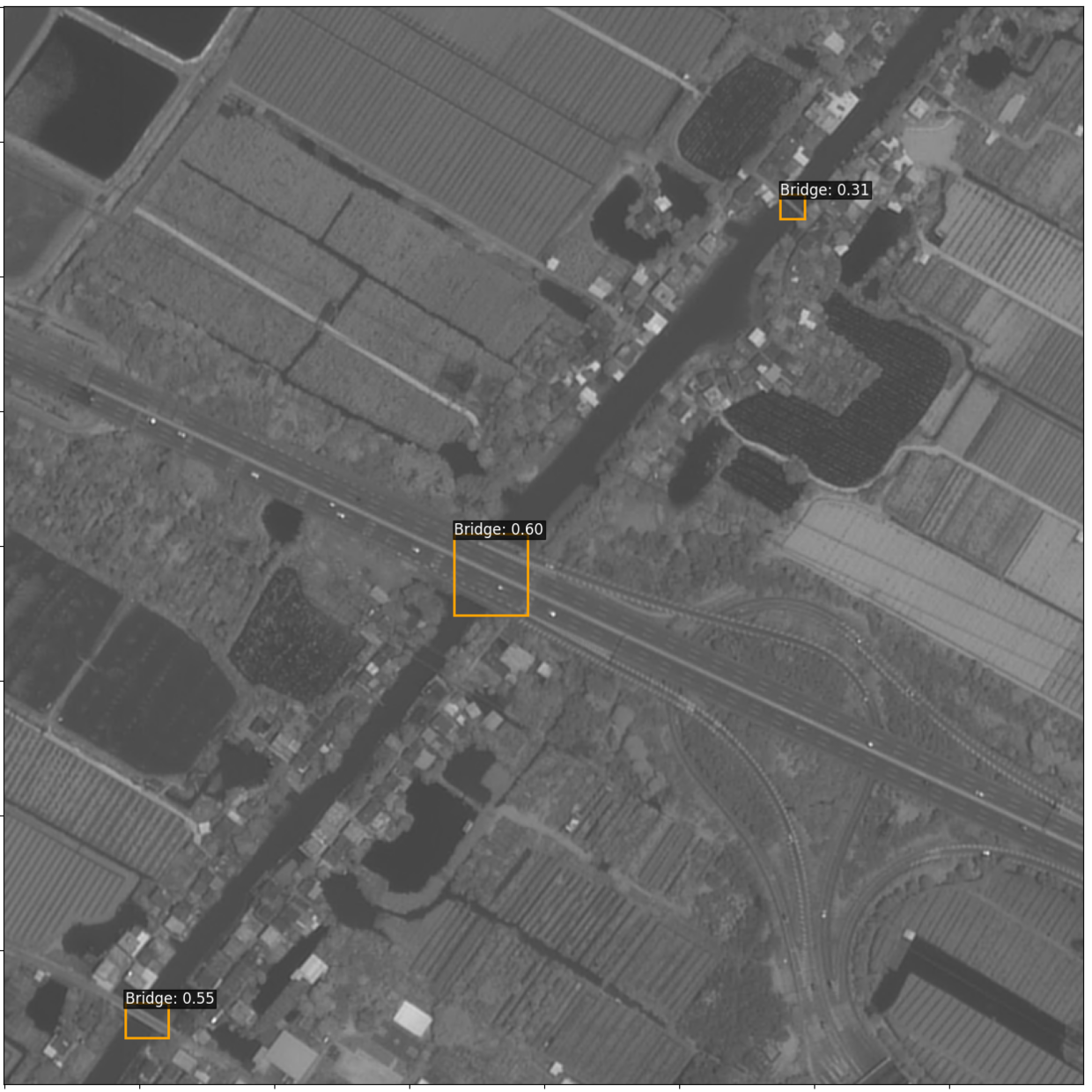}} &
 \adjustbox{margin=0pt}{\includegraphics[width=0.22\linewidth]{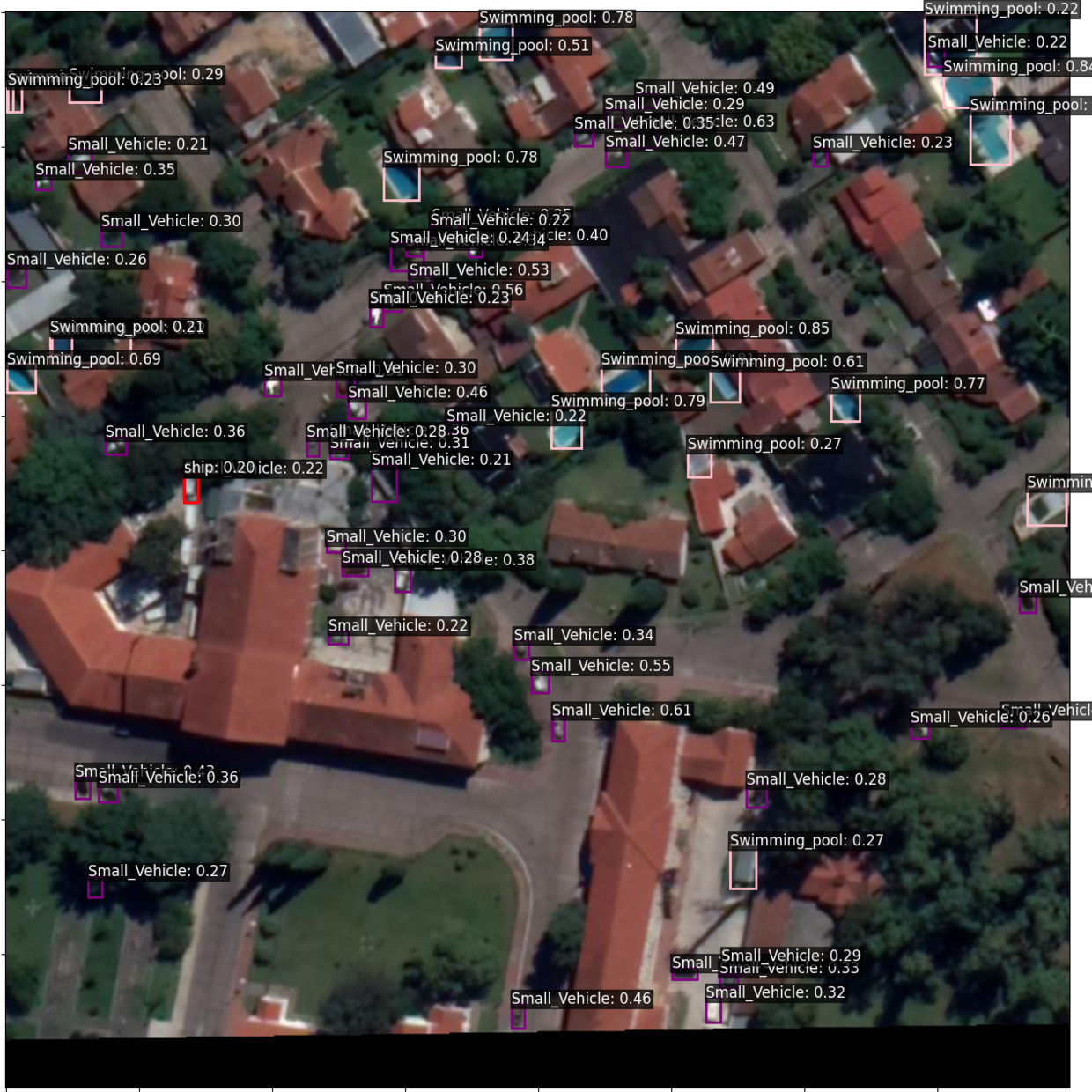}} \\
\end{tabular}
\caption{Some Qualitative Analysis on Our Model} 
\label{output}
\end{figure*}

This section delineates the outcomes of each experiment and provides insights into the implications of these results. It is important to note that all experiments involved singular modifications to the base model, which consists of LSKNet and DiffusionDet, making it the reference point for performance comparisons.

\subsection{Model Architectures and Their Impact}
Table~\ref{models} presents the results from varying model architectures. The integration of LSKNet as the backbone with DiffusionDet heads yielded the most favourable results, enhancing the mAP by approximately 1.8\% compared to the base Diffusion model with a Resnet backbone. Incorporating the Residual Connection improved performance by nearly 0.5\%, likely due to its ability to preserve initial image features before deeper feature extraction. However, adding a new block with randomly initialized parameters, as opposed to pre-trained weights like the rest of the blocks, resulted in reduced performance, indicating challenges in adapting these new weights within the overall backbone.
\setcounter{table}{0}
\begin{table}
\caption{Results of Different Model Architectures }

{\fontsize{9}{12}\selectfont 
\label{results}
    \centering
    \begin{tabular}{ | c | c | c | c | c | }
      \hline
      \thead{Experiment (Model)} &\thead{AP} & \thead{$AP_s$} & \thead{$AP_m$} & \thead{$AP_{l}$} \\
            \hline
      \makecell{Resnet + Faster RCNN}& \makecell{41.09}&  \makecell{26.52}  & \makecell{48.15}  &\makecell{ 53.13}   \\ 
        \hline
      \makecell{LSKNet + Faster RCNN}& \makecell{41.53}&  \makecell{27.46}  & \makecell{48.78}  &\makecell{ 53.40}   \\ 
        \hline
      \makecell{Resnet + DiffusionDet}& \makecell{40.72}&  \makecell{27.17}  & \makecell{46.67}  &\makecell{ 52.70}   \\ 
              \hline
      \makecell{Swin + DiffusionDet}& \makecell{36.53}&  \makecell{22.74}  & \makecell{44.56}  &\makecell{ 50.37}   \\ 
            \hline
      \makecell{\textbf {LSKNet + DiffusionDet*}}& \makecell{\textbf {42.48}}&  \makecell{\textbf {27.83}}  & \makecell{\textbf {49.08}}  &\makecell{\textbf  {55.23}}   \\ 
             \hline
      \makecell{\textbf {*Residual Connection}}& \makecell{\textbf {42.94}}&  \makecell{\textbf {28.11}}  & \makecell{\textbf {49.64}}  &\makecell{\textbf {55.48}}   \\ 
        \hline
      \makecell{*Added Block}& \makecell{40.21}&  \makecell{25.93}  & \makecell{46.48}  &\makecell{ 51.77}   \\ 
             \hline
    \end{tabular}\label{models}}
  \end{table}
  
\subsection{Effects of Loss and Activation Functions}
As shown in Table~\ref{loss}, experimenting with different loss and activation functions generally led to improvements in mAP, except for Weighted Focal Loss. Replacing GIOU with CIOU marginally increased accuracy but required more time to converge. Weighted Focal Loss underperformed, possibly due to imbalanced or excessive class weighting. Among the activation functions, Hardswish notably excelled in detecting smaller objects, a critical area for accuracy enhancement.

\begin{table}[h]
\caption{Results of Different Losses \& Activation Functions}
\centering
\fontsize{9}{12}\selectfont 
\begin{tabular}{ | c | c | c | c | c | }
  \hline
  \thead{Experiment (Model)} &\thead{AP} & \thead{$AP_s$} & \thead{$AP_m$} & \thead{$AP_{l}$} \\
  \hline
  \textbf{Smooth L1} & \textbf{43.40} & \textbf{28.57} & \textbf{49.87} & \textbf{55.74} \\ 
  \hline
  CIOU & 43.11 & 28.46 & 49.84 & 54.83 \\ 
  \hline
  Weighted Focal Loss & 41.51 & 27.29 & 49.13 & 53.99 \\ 
  \hline
  MISH & 42.76 & 27.67 & 50.17 & 54.40 \\  
  \hline
  \textbf{Hardswish} & \textbf{43.52} & \textbf{32.72} & \textbf{49.84} & \textbf{55.75} \\ 
  \hline
\end{tabular}
\label{loss}
\end{table}

\subsection{Hyperparameter Tuning and Its Effectiveness}
Table~\ref{hyper} reflects the outcomes of fine-tuning hyperparameters, demonstrating performance boosts from more customized settings. Altering aspect ratios appeared to benefit medium and large objects significantly. The most impactful individual modification was the increase in the number of proposals, elevating the model's mAP to 44.71\%. More images per batch also showed improvement, although it was sometimes limited by GPU memory constraints, occasionally leading to crashes. Both Soft NMS and augmentations exhibited potential for enhancing model performance.

\begin{table}
    \caption{Results of Different Configurations}
  {\fontsize{9}{12}\selectfont 
\label{results}
    \centering
    \begin{tabular}{ | c | c | c | c | c | }
      \hline
      \thead{Experiment (Model)} & \thead{AP} & \thead{$AP_s$} & \thead{$AP_m$} & \thead{$AP_{l}$} \\
            \hline
      \makecell{\textbf {Customized Aspect Ratio}}& \makecell{\textbf {44.58}}&  \makecell{\textbf {30.15}}  & \makecell{\textbf {51.66}}  &\makecell{\textbf {56.75}}   \\  
        \hline
      \makecell{\textbf {More Proposals}}& \makecell{\textbf {44.71}}&  \makecell{\textbf {29.82}}  & \makecell{\textbf {50.15}}  &\makecell{\textbf {57.22}}   \\  
                \hline
      \makecell{More Images Per Batch}& \makecell{ 44.42}&  \makecell{  29.24}  & \makecell{ 51.36}  &\makecell{57.97}   \\ 
            \hline
      \makecell{\textbf {Soft NMS}}& \makecell{\textbf {43.00}}&  \makecell{\textbf {27.93}}  & \makecell{\textbf {50.29}}  &\makecell{\textbf {54.11}}   \\ 
              \hline
      \makecell{\textbf {Augmentations}}& \makecell{\textbf {42.55}}&  \makecell{\textbf {27.94}}  & \makecell{\textbf {48.72}}  &\makecell{\textbf {54.40}}   \\ 
             \hline
    \end{tabular}\label{hyper}}
  \end{table}

\subsection{The Best Model and Its Superior Performance}
In pursuing the optimal model configuration, we combined various modifications individually, enhancing performance. The best model, highlighted in Table~\ref{best}, culminates these alterations. It integrates DiffusionDet with an LSKNet backbone, augmented with the extra residual connection, Hardswish activation function, Smooth L1, focal loss, and GIOU. Additionally, it incorporates the customized aspect ratio, increased proposals, and augmentations, with Soft NMS as the post-processing technique.

This best model achieved impressive results, notably a mean Average Precision (mAP) of \textit{45.7\%} on the test set, a significant accomplishment considering past literature and reports. The performance metrics for each class and across different object sizes in both test and validation sets are detailed in Table~\ref{best}.

\begin{table}[h]
\caption{Performance Metrics of the Best Model}
\centering
\fontsize{9}{12}\selectfont 
\begin{tabular}{ | c | c | c | c | c | }
  \hline
  \thead{Model} & \thead{AP} & \thead{$AP_s$} & \thead{$AP_m$} & \thead{$AP_{l}$} \\
  \hline
  \textbf{Best Model (Validation)} & \textbf{46.23} & \textbf{30.12} & \textbf{51.45} & \textbf{57.98} \\
    \hline
  \textbf{Best Model (Test)} & \textbf{45.70} & \textbf{29.87} & \textbf{51.05} & \textbf{57.31} \\
  \hline
\end{tabular}
\label{best}
\end{table}

Qualitatively, Figure~\ref{output} showcases the efficacy of our best model on random images from the dataset. The results demonstrate the model's capability to accurately detect objects, particularly smaller ones that were challenging for the base model.

\subsection{Observations}
Our rigorous experimentation has resulted in a model that notably enhances object detection in aerial imagery, establishing new performance benchmarks. Remarkably, our model, constrained to a single GPU and limited iterations, demonstrates competitive efficiency compared to state-of-the-art models \cite{isaid} that utilized 8 GPUs and 180,000 iterations. This achievement highlights our model's energy efficiency and potential for scalability. The focused modifications in model architecture, loss functions, activation functions, and hyperparameters, even within hardware limitations, have significantly advanced aerial image analysis. These results indicate that our model could achieve further substantial improvements in the field with access to more robust computational resources.

\section{Conclusion}
In this research, we have introduced a suite of innovative enhancements that substantially elevate the good results in aerial image analysis. Our methodology entailed the development of a robust and sophisticated backbone, integrating large kernel convolutions, spatial kernel selection, and feature pyramid networks. This backbone was further augmented with adapted diffusion models tailored specifically for the complexities of aerial imaging, thereby improving object detection and classification.

The architectural advancements in our proposed model yielded a more powerful and efficient tool for aerial image analysis. We conducted extensive investigations into various activation functions, ultimately identifying the most effective option for our specific application. To address the prevalent class imbalance issue, we devised a weighted focal loss function and explored the adaptation of additional loss functions for box regression.

Our exhaustive examination and fine-tuning of hyperparameters and post-processing methods culminated in an optimized model with notable improvements. These efforts culminated in a significant increase in mean Average Precision (mAP), achieving a mAP of 45.7\% on the test dataset. This comprehensive approach marks a significant stride forward in enhancing the accuracy and robustness of aerial image analysis.

\section{Limitations \& Future Work}
This study encountered certain limitations that impacted our ability to achieve even higher performance metrics. A primary constraint was the GPU memory capacity, which restricted our ability to increase the number of images per batch. This limitation was particularly notable, as our best model demonstrated potential for accuracy improvement with larger batch size, but we often faced memory overflow or system crashes.

Another challenge was the unavailability of LSKNet pre-trained weights for the COCO dataset\cite{lin2014microsoft}. Our experiments revealed that COCO pre-trained weights generally outperform those from ImageNet for object detection tasks. We partially mitigated this by using COCO pre-trained weights for the DiffusionDet heads, but this led to inconsistent weight distributions between the backbone and the head. Despite efforts to fine-tune the model on COCO, time constraints and frequent GPU crashes limited our progress.

Looking forward, our future work aims to address these limitations. Access to higher-capacity GPUs and extended training time on the COCO dataset are among our primary objectives. We anticipate that enabling a larger number of images per batch could significantly enhance model performance, further advancing the field of aerial image analysis.

{\small
\bibliographystyle{ieee_fullname}
\bibliography{egbib}
}


    
    

\end{document}